\documentclass[10pt,twocolumn,letterpaper]{article}
\usepackage{tikz}
\usepackage{cvpr}
\usepackage{times}
\usepackage{epsfig}
\usepackage{graphicx}
\usepackage{amsmath}
\usepackage{amssymb}
\usepackage[noend]{algpseudocode}
\usepackage{algorithm}

\usepackage{pgfplots}
\pgfplotsset{compat=newest}
\usetikzlibrary{plotmarks}
\usepackage{booktabs} % Nice tables
\usepackage{multirow}
\usepackage{paralist, tabularx}

\usepackage[pagebackref=true,breaklinks=true,colorlinks,bookmarks=false]{hyperref}

\cvprfinalcopy % *** Uncomment this line for the final submission

\def\cvprPaperID{436} % *** Enter the CVPR Paper ID here

\ifcvprfinal\fi
\begin{document}

%%%%%%%%% TITLE
\title{Oracle MCG: A first peek into COCO Detection Challenges\vspace{-2mm}}

\author{Jordi Pont-Tuset$^{1}\qquad\qquad$ Pablo Arbel\'aez$^{2}\qquad\qquad$ Luc Van Gool$^1$\\[2mm]\and
$^1$ETH Zurich, Switzerland\\[-1mm]{\tt\small \{jponttuset,vangool\}@vision.ee.ethz.ch}
\and
$^2$Universidad de los Andes, Colombia\\[-1mm]
{\tt\small pa.arbelaez@uniandes.edu.co}
}

\maketitle
Microsoft COCO~\cite{Lin2014a} is a new annotated database in computer vision consisting of more than 200.000 images.
There are currently more than one million annotated objects from 80 categories, with fully segmented masks.
With respect to Pascal~\cite{Everingham2012}, the previous available dataset with semantic segmentation annotations, COCO has four times the number of categories and two orders of magnitude more images and annotated objects.

In this context, the challenges for object detection in COCO have recently been 
presented.
In a nutshell, competing methods should provide a list of detections on each image
in the form of bounding boxes or segmentation masks.
Each detection should have associated one of the 80 categories and a confidence score.

These new challenges on a new dataset are \textit{unchartered territory} for 
researchers used to work on Pascal for some years.
This work's main aim is to set a reference point in the challenges to get a
first grasp of which results to expect.

To do so, we provide \textbf{Oracle MCG}, a hypothetical detector consisting of an oracle picking the best object proposal of a state-of-the-art technique on all annotated objects.
This result could be understood as the upper-bound result one could get by restricting themselves to selecting object proposals without any further refinement.

In particular, we use the publicly-available pre-computed results given by
MCG~\cite{Pont-Tuset2015} on the validation set of COCO (also available for the test set).
On average, this method produces 5075 proposals per image.
We then overlap all proposals on each image against all annotated objects 
and pick the best one for each of them.
Each of these best proposals is associated to the annotated category and assigned 
a constant score of 1.

The results obtained, as given by the COCO official evaluation software, are shown in Table~\ref{oracle_boxes} for bounding boxes
and in Table~\ref{oracle_segm} for segmentation masks.
The highlighted row is the one that will define the ranking of the competition.
Below some observations about the results.

\begin{table}[h]
%\centering
%\scalebox{0.7}{\begin{tabular}{c@{\hspace{2.5mm}}c@{\hspace{2.5mm}}c@{\hspace{2.5mm}}c@{\hspace{2.5mm}}c@{\hspace{2.5mm}}c@{\hspace{2.5mm}}c|@{\hspace{2.5mm}}c@{\hspace{2.5mm}}c@{\hspace{2.5mm}}c@{\hspace{2.5mm}}c@{\hspace{2.5mm}}c@{\hspace{2.5mm}}c}
%\cmidrule[\heavyrulewidth](l{-2pt}){2-13}
% & \rotatebox{90}{AP Standard} & \rotatebox{90}{AP IoU=0.5} & \rotatebox{90}{AP IoU=0.75} & \rotatebox{90}{AP Small} & \rotatebox{90}{AP Medium} & \rotatebox{90}{AP Large} &  \rotatebox{90}{AR Standard} & \rotatebox{90}{AR MaxDet=1} & \rotatebox{90}{AR MaxDet=10} & \rotatebox{90}{AR Small} & \rotatebox{90}{AR Medium} & \rotatebox{90}{AR Large}\\
%\midrule
%BBox & \\
%Segm & \textbf{.292} & .605 & .253 & .122 & .341 & .533 & .299 & .439 & .445 & .240 & .494 & .671\\
%\bottomrule
%\end{tabular}}
%\vspace{1mm}
\noindent\scalebox{0.515}{\textbf{\texttt{Average Precision (AP) @[ IoU=0.50:0.95 | area=\ \ \ all | maxDets=100 ] = 0.317}}}\\[-2mm]
 \scalebox{0.515}{\texttt{Average Precision (AP) @[ IoU=0.50\ \ \ \ \ \ | area=\ \ \ all | maxDets=100 ] = 0.599}}\\[-2mm]
 \scalebox{0.515}{\texttt{Average Precision (AP) @[ IoU=0.75\ \ \ \ \ \ | area=\ \ \ all | maxDets=100 ] = 0.295}}\\[-2mm]
 \scalebox{0.515}{\texttt{Average Precision (AP) @[ IoU=0.50:0.95 | area= small | maxDets=100 ] = 0.143}}\\[-2mm]
 \scalebox{0.515}{\texttt{Average Precision (AP) @[ IoU=0.50:0.95 | area=medium | maxDets=100 ] = 0.369}}\\[-2mm]
 \scalebox{0.515}{\texttt{Average Precision (AP) @[ IoU=0.50:0.95 | area= large | maxDets=100 ] = 0.553}}\\[-2mm]
 \scalebox{0.515}{\texttt{Average Recall\ \ \ \ (AR) @[ IoU=0.50:0.95 | area=\ \ \ all | maxDets=\ \ 1 ] = 0.319}}\\[-2mm]
 \scalebox{0.515}{\texttt{Average Recall\ \ \ \ (AR) @[ IoU=0.50:0.95 | area=\ \ \ all | maxDets= 10 ] = 0.476}}\\[-2mm]
 \scalebox{0.515}{\texttt{Average Recall\ \ \ \ (AR) @[ IoU=0.50:0.95 | area=\ \ \ all | maxDets=100 ] = 0.483}}\\[-2mm]
 \scalebox{0.515}{\texttt{Average Recall\ \ \ \ (AR) @[ IoU=0.50:0.95 | area= small | maxDets=100 ] = 0.274}}\\[-2mm]
 \scalebox{0.515}{\texttt{Average Recall\ \ \ \ (AR) @[ IoU=0.50:0.95 | area=medium | maxDets=100 ] = 0.536}}\\[-2mm]
 \scalebox{0.515}{\texttt{Average Recall\ \ \ \ (AR) @[ IoU=0.50:0.95 | area= large | maxDets=100 ] = 0.708}}\\[-3mm]
\caption{Oracle MCG COCO evaluation as bounding boxes}
\label{oracle_boxes}
\end{table}

\begin{table}[h]
%\centering
%\scalebox{0.7}{\begin{tabular}{c@{\hspace{2.5mm}}c@{\hspace{2.5mm}}c@{\hspace{2.5mm}}c@{\hspace{2.5mm}}c@{\hspace{2.5mm}}c@{\hspace{2.5mm}}c|@{\hspace{2.5mm}}c@{\hspace{2.5mm}}c@{\hspace{2.5mm}}c@{\hspace{2.5mm}}c@{\hspace{2.5mm}}c@{\hspace{2.5mm}}c}
%\cmidrule[\heavyrulewidth](l{-2pt}){2-13}
% & \rotatebox{90}{AP Standard} & \rotatebox{90}{AP IoU=0.5} & \rotatebox{90}{AP IoU=0.75} & \rotatebox{90}{AP Small} & \rotatebox{90}{AP Medium} & \rotatebox{90}{AP Large} &  \rotatebox{90}{AR Standard} & \rotatebox{90}{AR MaxDet=1} & \rotatebox{90}{AR MaxDet=10} & \rotatebox{90}{AR Small} & \rotatebox{90}{AR Medium} & \rotatebox{90}{AR Large}\\
%\midrule
%BBox & \\
%Segm & \textbf{.292} & .605 & .253 & .122 & .341 & .533 & .299 & .439 & .445 & .240 & .494 & .671\\
%\bottomrule
%\end{tabular}}
%\vspace{1mm}
\noindent\scalebox{0.515}{\textbf{\texttt{Average Precision (AP) @[ IoU=0.50:0.95 | area=\ \ \ all | maxDets=100 ] = 0.292}}}\\[-2mm]
 \scalebox{0.515}{\texttt{Average Precision (AP) @[ IoU=0.50\ \ \ \ \ \ | area=\ \ \ all | maxDets=100 ] = 0.605}}\\[-2mm]
 \scalebox{0.515}{\texttt{Average Precision (AP) @[ IoU=0.75\ \ \ \ \ \ | area=\ \ \ all | maxDets=100 ] = 0.253}}\\[-2mm]
 \scalebox{0.515}{\texttt{Average Precision (AP) @[ IoU=0.50:0.95 | area= small | maxDets=100 ] = 0.122}}\\[-2mm]
 \scalebox{0.515}{\texttt{Average Precision (AP) @[ IoU=0.50:0.95 | area=medium | maxDets=100 ] = 0.341}}\\[-2mm]
 \scalebox{0.515}{\texttt{Average Precision (AP) @[ IoU=0.50:0.95 | area= large | maxDets=100 ] = 0.533}}\\[-2mm]
 \scalebox{0.515}{\texttt{Average Recall\ \ \ \ (AR) @[ IoU=0.50:0.95 | area=\ \ \ all | maxDets=\ \ 1 ] = 0.299}}\\[-2mm]
 \scalebox{0.515}{\texttt{Average Recall\ \ \ \ (AR) @[ IoU=0.50:0.95 | area=\ \ \ all | maxDets= 10 ] = 0.439}}\\[-2mm]
 \scalebox{0.515}{\texttt{Average Recall\ \ \ \ (AR) @[ IoU=0.50:0.95 | area=\ \ \ all | maxDets=100 ] = 0.445}}\\[-2mm]
 \scalebox{0.515}{\texttt{Average Recall\ \ \ \ (AR) @[ IoU=0.50:0.95 | area= small | maxDets=100 ] = 0.240}}\\[-2mm]
 \scalebox{0.515}{\texttt{Average Recall\ \ \ \ (AR) @[ IoU=0.50:0.95 | area=medium | maxDets=100 ] = 0.494}}\\[-2mm]
 \scalebox{0.515}{\texttt{Average Recall\ \ \ \ (AR) @[ IoU=0.50:0.95 | area= large | maxDets=100 ] = 0.671}}\\[-3mm]
\caption{Oracle MCG COCO evaluation as segmented masks}
\label{oracle_segm}
\vspace{-2mm}
\end{table}

\begin{itemize}
\item COCO detection is a very challenging competition, given that an oracle selecting the best out of a set of 5000 proposals
only reaches a performance around 0.3.
\item Although detecting segmented masks is in theory much more challenging than detecting at bounding box level, the results on bounding boxes are not extremely better than for segmentation.
\item Small objects are very relevant in COCO. While the result for large objects is 0.533, this drops to 0.122 in small objects and goes up only to 0.292 overall.
\item The ranking measure gives a strong weight to the very precise delineation of objects by sweeping IoU thresholds up to 0.95.
The results for the threshold at 0.5 is 0.605 compared to 0.292.
\end{itemize}

{\small
\bibliographystyle{ieee}
\bibliography{PTAVG_MCG_oracle}
}

\end{document}